\documentclass[tinyml]{acmart}

\AtBeginDocument{%
  \providecommand\BibTeX{{%
    \normalfont B\kern-0.5em{\scshape i\kern-0.25em b}\kern-0.8em\TeX}}}

\setcopyright{rightsretained}
\copyrightyear{2021}
\acmYear{2021}

\usepackage{soul}
\usepackage{amsmath}
\usepackage{tabularx}
\usepackage{multirow}
\usepackage{times}
\usepackage{epsfig}
\usepackage{graphicx}
\usepackage{makecell}
\begin{document}

\title{TTVOS: Lightweight Video Object Segmentation with Adaptive Template Attention Module and Temporal Consistency Loss}
\renewcommand{\shorttitle}{TTVOS: Lightweight Video Object Segmentation }
% \affiliation{%

\author{Hyojin Park}
% \authornotemark[1] %\footnote{This work was done during Hyojin's internship at Facebook}
% \authornote{This work was done during Hyojin's internship at Facebook.}
\email{wolfrun@snu.ac.kr}
\affiliation{%
  \institution{Seoul National University}
  \city{Seoul}
  \country{Republic of Korea}
}

\author{Ganesh Venkatesh}
\email{gven@fb.com}
\affiliation{%
  \institution{Facebook, Inc.}
  \state{California}
  \country{USA}
}
\author{Nojun Kwak}
\email{ nojunk@snu.ac.kr}
\affiliation{%
  \institution{Seoul National University}
  \city{Seoul}
  \country{Republic of Korea}
}
\renewcommand{\shortauthors}{H. Park, G. Venkatesh, and N. Kwak}


\begin{abstract}
 
\textit{Semi-supervised video object segmentation} (semi-VOS) is widely used in many applications.
This task is tracking class-agnostic objects from a given target mask.
For doing this, various approaches have been developed based on online-learning, memory networks, and optical flow.
These methods show high accuracy but are hard to be utilized in real-world applications due to slow inference time and tremendous complexity.
To resolve this problem, template matching methods are devised for fast processing speed but sacrificing lots of performance in previous models.
We introduce a novel semi-VOS model based on a template matching method and a temporal consistency loss to reduce the performance gap from heavy models while expediting inference time a lot.
Our template matching method consists of short-term and long-term matching.
The short-term matching enhances target object localization, while long-term matching improves fine details and handles object shape-changing through the newly proposed adaptive template attention module.
However, the long-term matching causes error-propagation due to the inflow of the past estimated results when updating the template.
To mitigate this problem, we also propose a temporal consistency loss for better temporal coherence between neighboring frames by adopting the concept of a transition matrix.
Our model obtains $79.5\%$  $J\&F$ score at the speed of 73.8 FPS on the DAVIS16 benchmark.
The code is available in https://github.com/HYOJINPARK/TTVOS.
\end{abstract}

\keywords{Semi-supervised video segmentation, video tracking, video object segmentation}

\maketitle

\section{Introduction}
\label{sec:intro}

\begin{figure}[t]
\centering
% \fbox{\rule{0pt}{2.5in} \rule{.9\linewidth}{0pt}}
    \includegraphics[width=0.98\linewidth]{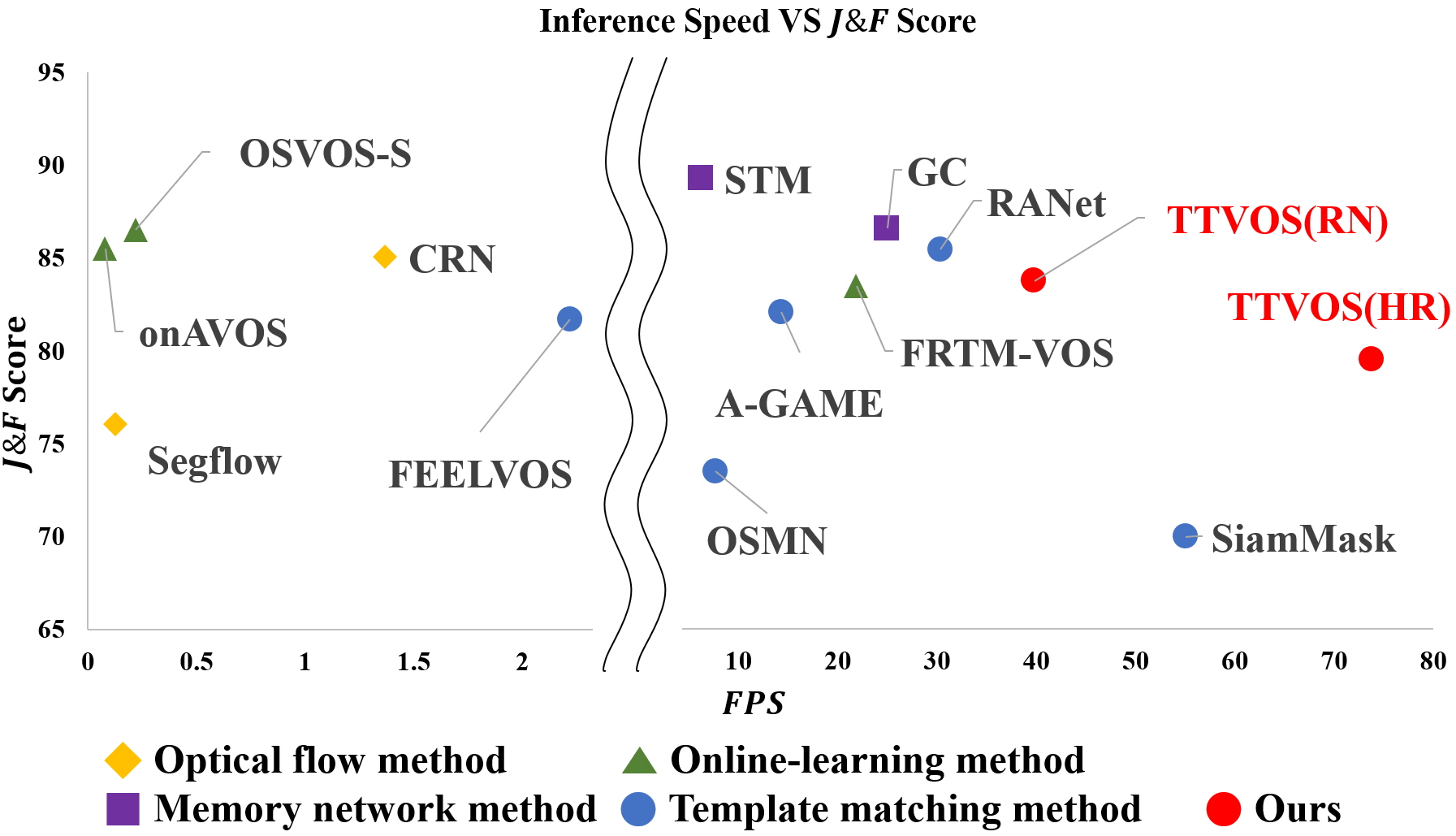}  
   \caption{The speed (FPS) vs accuracy ($J\&F$ score) on the DAVIS2016 validation set. 
  Our proposed TTVOS achieves high accuracy with small complexity.
   HR/RN respectively denotes HRNet/ResNet50 for the backbone network.}
\label{fig:comparison}
\end{figure}

Video object segmentation (VOS) is essential in many applications such as autonomous driving, video editing, and surveillance system. 
In this paper, we focus on a \textit{semi-supervised video object segmentation} (semi-VOS) task, which is to find a target in a pixel-wise resolution from a given annotated mask for the first frame. 

For accurate tracking, many models have been developed, but it is hard to use the models in real-world environment due to tremendous computation.
For example, one of popular method, online-learning, fine-tunes model parameters using the first frame image and the corresponding ground truth mask~\cite{robinson2020learning,maninis2018video,perazzi2017learning,Cae+17}.
This strategy makes the model more specialize in each video input, but, it requires additional time and memory for fine-tuning.
Memory network method achieves high accuracy than any other approaches.
The model stacks multiple target memories and matches the current frame with the memories.
Therefore, the inference time and the required memories increase in proportion to the number of frames. 
To solve these problems, GC \cite{li2020fast} conducted weighted-average to the multiple memories at each time frame for generating one global context memory.
However, it still needs an additional feature extraction step for updating the memory from the current estimated mask and the image.
Also, we believe that it is not enough to directly comprehend spatial information since the size of global context memory much smaller than original spatial resolution size.

For increasing consistency of masks across frames, optical flow is one of the popular methods in low-level vision which has been applied in diverse video applications.
In a video segmentation task, it re-aligns a given mask or features by computing pixel-wise trajectories or movements of objects as an additional clue \cite{lin2020flow,wang2018semi,hu2018motion,cheng2017segflow}. 
However, it is too demanding to compute exact flow vectors which contain excessive information for the segmentation task.
For example, if we know the binary information of whether a pixel is changed into the foreground or background, we do not need an exact flow vector of each pixel.

The aforementioned methods have increased accuracy a lot, but they require heavy inference time and memory.
The template matching approach resolves this problem by designing a target template from a given image and annotation.
However, the accuracy is lower compared to other models because the matching method is too simple, and the template is hard to handle object shape variation

In this paper, we propose an adaptive template matching method and a novel temporal consistency loss for semi-VOS.
Our contributions can be summarized as follows:
1) We propose a new lightweight VOS model based on template matching method by combining short-term and long-term matching to achieve fast inference time and to reduce the accuracy gap from heavy and complex models.
More specifically, in short-term matching, we compare the current frame’s feature with the information from the previous frame for localization.
In long-term matching, we devise an adaptive template for generating an accurate mask.
2) We introduce a novel adaptive template motivated from GC for managing shape variation of target objects.
Our adaptive template is updated from the current estimated mask without re-extracting features and occupying additional memory.
3) To improve performance of model, we propose a new temporal consistency loss for mitigating the error propagation problem which is one of the main reasons for accuracy degradation caused by inflow of the past estimated results.
To the best of our knowledge, this work is the first to apply the concept of consistency loss for the semi-VOS task without optical flow.
Our model generates a transition matrix to encourage the correction of the incorrectly estimated pixels from the previous frame and preventing their propagation to future frames.
Our model achieves $79.5\%$ $J\&F$ score at the speed of 73.8 FPS on the DAVIS16 benchmark (See Fig. \ref{fig:comparison}).
We also verified the efficacy of the temporal consistency loss by applying it to other models and showing increased performance.

\section{Related Work}
\label{sec:related}

\noindent
\textbf{Online-learning: }
The online-learning method is training the model with new data in inference stage \cite{ijcai2018-369,zhou2012online,kivinen2004online}.
In the semi-VOS task, model parameters are fine-tuned in the inference stage with a given input image and a corresponding target mask.
Therefore, the model is specialized for tracking the target \cite{maninis2018video,perazzi2017learning,Cae+17}. 
However, fine-tuning causes additional latency in inference time.
\cite{robinson2020learning} resolved this issue by dividing the model into two sub-networks. 
One is a lightweight network that is fine-tuned in the inference stage for making a coarse score map. The other is a heavy segmentation network without the need for fine-tuning.
This network enables fast optimization and relieves the burden of online-learning.

\noindent
\textbf{Memory network: }
The memory network constructs external memory representing various properties of the target.
It was devised for handling long-term sequential tasks in the natural language processing (NLP) domain, such as the QA task~\cite{kim2019progressive,sukhbaatar2015end,weston2014memory}.
STM \cite{oh2019video} adopted this idea for the semi-VOS task by a new definition of key and value.
The \textit{key} encodes visual semantic clue for matching and the \textit{value} stores detailed information for making the mask.
However, it requires lots of resources because the amount of memory is increased over time.
Furthermore, the size of memory is the square of the resolution of an input feature map.
To lower this huge complexity, GC \cite{li2020fast} does not stack memory at each time frame, but accumulate them into one, which is also of a smaller size than a unit memory of STM. 
They does not make a $(hw\times hw)$ memory like \cite{zhu2019asymmetric,wang2018non} but a $(c_{key}\times c_{val})$ memory\footnote{$h$ and $w$ are the height and the width of an input feature map for constructing memory, and $c_{key}$ and $c_{val}$ are the number channels for the key and value feature maps.} as similar channel attention module.

\noindent
\textbf{Template matching: }
Template matching is one of the traditional method in the tracking task.
It generates a template and calculates similarity with input as a matching operation.
Most works follow the siamese network~\cite{bertinetto2016fully} approach which makes target template feature and a feature map of a given image from same network for matching operation.
RANet \cite{wang2019ranet} applied a racking system to the matching process between multiple templates and input for extracting reliable results. 
FEELVOS \cite{voigtlaender2019feelvos} calculated distance map by local and global matching for better robustness.
SiamMask \cite{wang2019fast} used a depth-wise operation for fast matching and makes a template from a bounding box annotation without accurate annotated mask of a target.
A-GAME \cite{johnander2019generative} proposed other method which designed a target distribution by a mixture of Gaussian in an embedding space.
It predicted posterior class probabilities for matching.

\noindent
\textbf{Optical flow: }
Optical flow estimates flow vectors of moving objects and widely used in many video applications \cite{khoreva2017lucid,dutt2017fusionseg,tsai2016video,sevilla2016optical}.
In the semi-VOS task, it is provided to re-aligns the given mask or features from previous to current frame for encouraging temporal consistency. 
Segflow \cite{cheng2017segflow} designed two branches, each for image segmentation and optical flow.
The outputs of both branches are combined together to estimate the target masks.
Similarly, FAVOS \cite{lin2020flow} and CRN \cite{hu2018motion} refined a rough segmentation mask by optical flow.

\noindent
\textbf{Consistency Loss: }
Consistency loss is widely used for improving performance or robustness in semi-supervised learning.
It injects perturbation into input to learn generation of stable output \cite{miyato2018virtual,zhu2017unpaired}.
In VOS, consistency usually means temporal coherence between neighboring frames by additional clue from optical flow \cite{Tsai_2016_CVPR,volz2011modeling,weickert2001variational}.

\begin{figure*}[t]
\centering
   \includegraphics[width=0.80\textwidth]{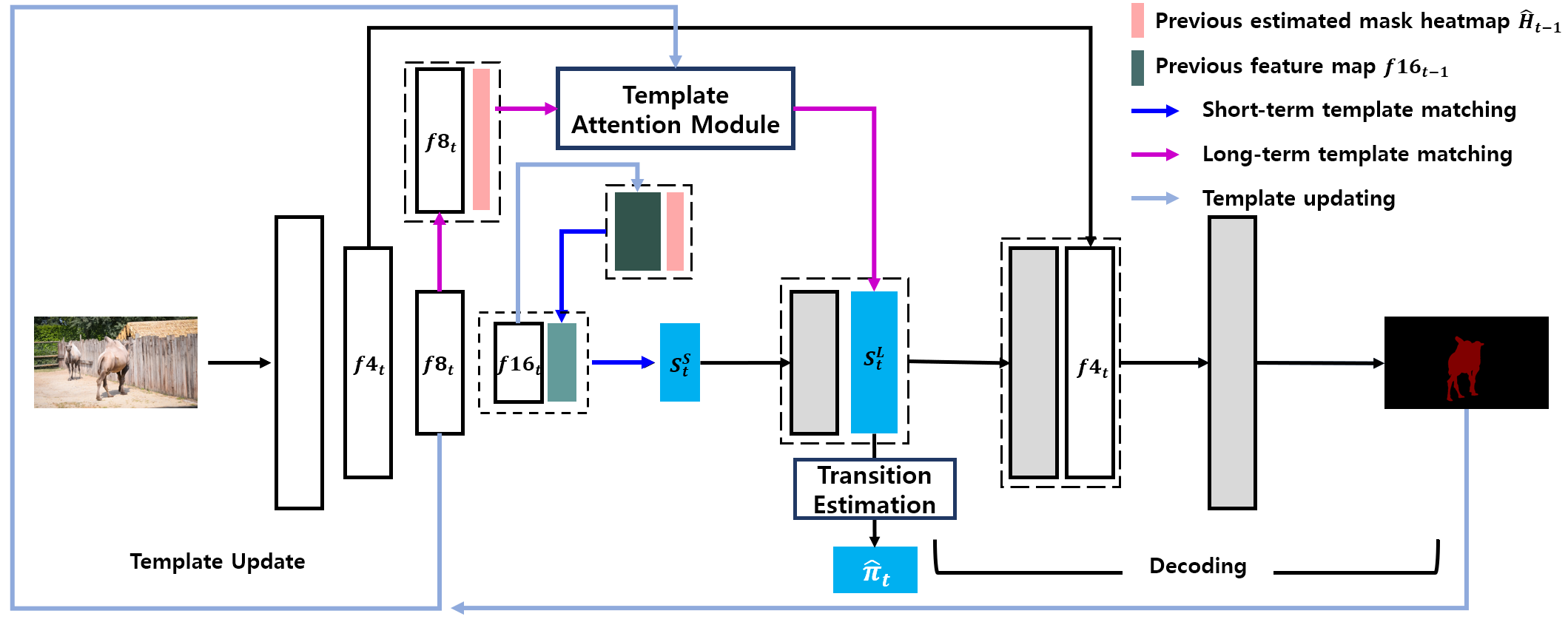}
   \caption{The overall architecture of TTVOS. A backbone feature is shared in all the processes of TTVOS for efficiency. 
   There are two types of template matching (short-term and long-term), decoding and template update stages in our model. The transition matrix $\hat{\pi}_t$ is computed only in the training phase for enhancing temporal coherence.}
\label{fig:Network}
\end{figure*}
%%-------------------------Fig of arch -------------------------------------

\section{Method}
\label{sec:method}
In this section, we present our semi-VOS model.
Section \ref{subsec:network} introduces the whole model architecture and how to manage multi-object VOS.
Section \ref{subsec:template} explains the details of template attention module for producing a  similarity map by long-term matching.
We also describe how to update the long-term template.
Finally, Section \ref{subsec:TCloss} demonstrates our temporal consistency loss and how to define new ground truth for mitigating error propagation between neighboring frames. 

\subsection{Overall TTVOS Architecture}
\label{subsec:network}

We propose a new architecture for VOS as shown in Fig. \ref{fig:Network}.
Our TTVOS consists of feature extraction, template matching, decoding, and template update stages.
The template matching is composed of a short-term matching and a long-term matching.
The short-term matching enhances localization property by using previous information.
This uses a small feature map for producing a coarse segmentation map.
However, this incurs two problems: 
1) Utilizing only the information of the previous frame causes the output masks overly dependent on previous results.
2) This can not handle shape-changing nor manifest detailed target shape due to a small feature map.
To resolve these problems, we propose long-term matching as an adaptive template matching method.
This template is initialized from the given first frame condition and updated at each frame.
Therefore, it can consider the whole frames and track gradually changing objects. 
This module uses a larger feature map for getting more hihg-resolution information for generating accurate masks.
After then, our model executes decoding and updates each templates.

A backbone extracts feature maps $fN_t$ from the current frame, where $fN_t$ denotes a feature map at frame $t$ with an $1/N$-sized width and height compared to the input. 
Short-term matching uses a small feature map $f16_t$ and the previous frame information for target localization:
$f16_{t-1}$ is concatenated with a previous mask heatmap $\hat{H}_{t-1}$, which consists of two channels containing the probability of background and foreground respectively.
After then, this concatenated feature map is forwarded by several convolution layers for embedding short-term template.
The short-term template is matched with current feature map $f16_t$ to calculating short-term similarity map $S_t^S$ for localization.
In the long-term template matching stage, $f8_t$ is concatenated with the previous mask heatmap for comparing with the adaptive template to produce a long-term similarity map $S_t^L$ in the template attention module as detailed in Section \ref{subsec:template}.
Finally the long-term similarity map and the upsampled short-term similarity map is concatenated for decoding stage.

In decoding stage, $f4_t$ is added to the concatenated similarity map for a more accurate mask.
We use ConvTranspose for upsampling and use PixelShuffle \cite{shi2016real} in the final upsampling stage to prevent the grid-effect.
After target mask estimation, $f16_t$ and $\hat{H}_{t}$ are used for updating next short-term template matching, and $f8_t$ and $\hat{H}_{t}$ are utilized for next long-term template matching.
At only training time, the long-term similarity map $S_t^L$ estimates a transition matrix $\hat{\pi_t}$ for temporal consistency loss.
The loss guides the model to improve consistency of masks between neighboring frames, and the detailed is explained in Section \ref{subsec:TCloss}.
All the backbone features are also shared in the multi-object case, but the stages of two template matching and decoding are conducted separately for each object. 
Therefore, each object's heatpmap always has two channels for the probability of background and foreground. 
At inference time, all the heatmaps are combined by the soft aggregation method \cite{cho2020crvos,johnander2019generative}.

\begin{figure*}[t]
\centering
\begin{tabular}{ccc}

    \includegraphics[width=0.25\linewidth]{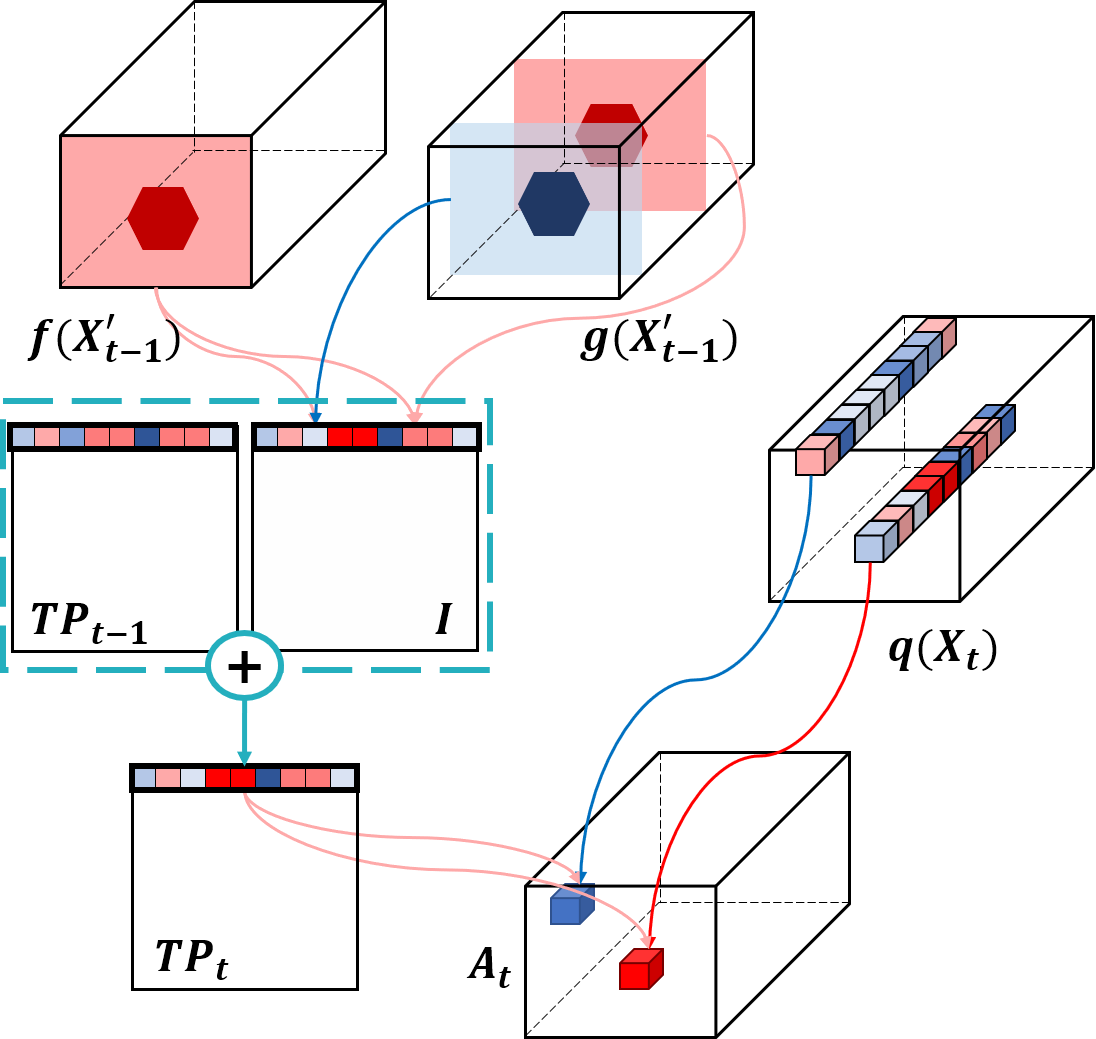}&
    \includegraphics[width=0.50\linewidth]{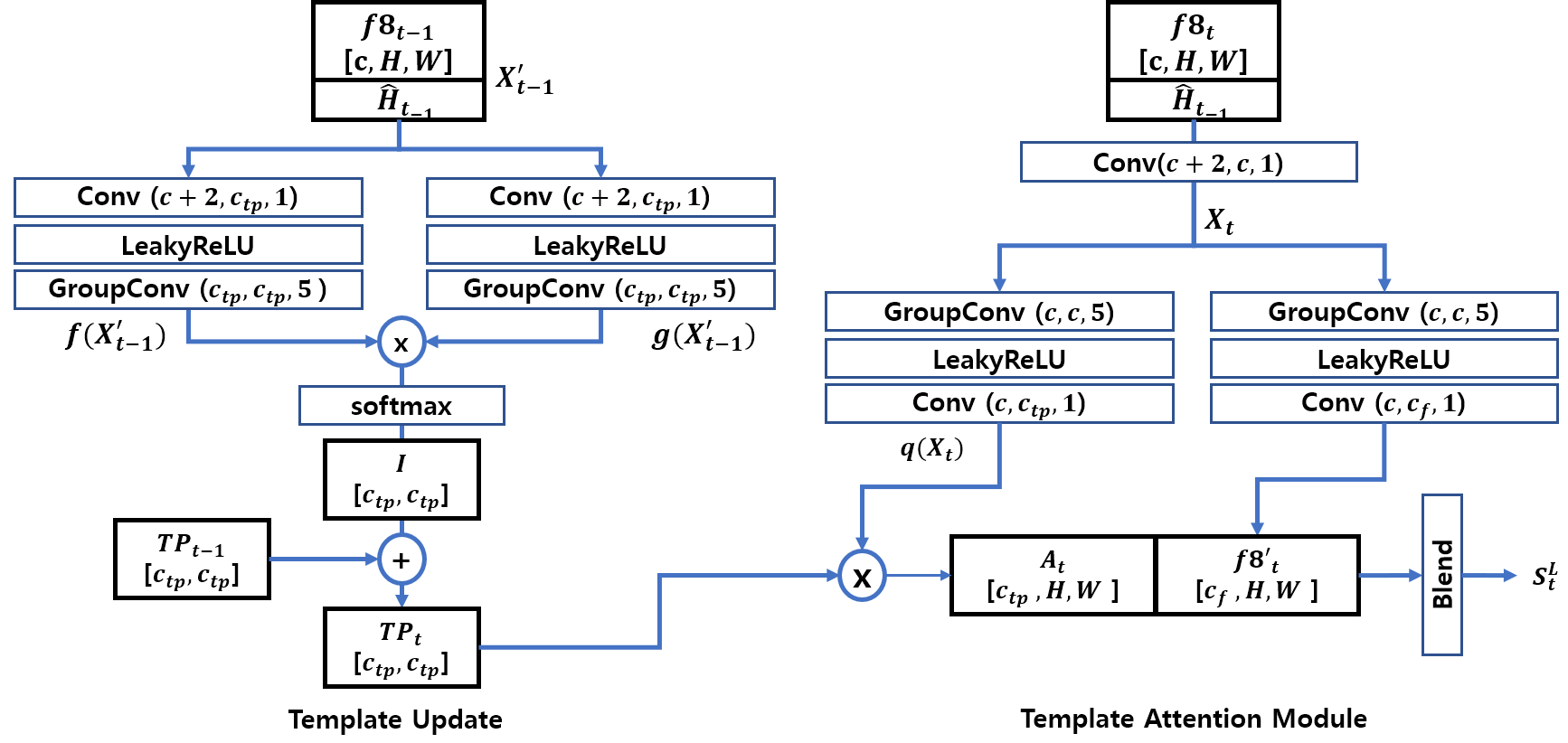}  \\
    (a)&(b)  \\
        \end{tabular}%

  \caption{(a) Process in a template attention module.
  Here, a red (blue) color means a high (low) similarity between two information.  
  The size of $f(X'_{t-1})$ and $g(X_{t-1})$ is $c_{tp}\times HW$, but we draw feature maps as $c_{tp}\times H\times W$ for the sake of convenient understanding. 
  (b) The detailed structure of a template attention module and a template update. 
   An operation (a,b,c) denotes the input channel, output channel, and kernel size of convolution operation, respectively.}
\label{fig:template}
\end{figure*}

\subsection{Template Attention Module}
\label{subsec:template}

We conjecture that pixels inside a target object have a distinct embedding vector distinguished from non-target object pixels.
Our model is designed to find this vector by self-attention while suppressing the irrelevant information of the target object.
Each current embedding vector updates a previous long-term template by weighted-average at each frame.
The proposed module generates a attention map  $A_t$ by using the template to make accurate mask as shown in Fig. \ref{fig:template}.

For constructing the current embedding vector, the backbone feature $f8_{t-1}$ and the previous estimated mask heatmap $\hat{H}_{t-1}$ are concatenated to suppress information far from the target object.
In Fig. \ref{fig:template}, the concatenated feature map is denoted as $X'_{t-1}$.
$X'_{t-1}$ is forwarded to two separate branches$f(\cdot)$ and $g(\cdot)$, making $f(X'_{t-1}), g(X'_{t-1}) \in \mathbb{R}^{c_{tp}\times H\times W}$.
After then, the feature maps are reshaped to $c_{tp}\times HW$ and producted to generate an embedding matrix $I$ as follows:
\begin{equation}
    \label{eq:attention}
     I =\sigma ( f(X'_{t-1}) \times g(X'_{t-1})^T) \in \mathbb{R}^{c_{tp} \times c_{tp}}.
\end{equation}
Here, $\sigma$ is a softmax function applied row-wise.
$I_{i,j}$ is the $(i,j)$ element of $I$, and it is generated by dot-product along $HW$ direction.
In other words, it represents an $i$th channel's view about $j$th channel information by aggregating all information in the $HW$ plane while preventing the inflow of irrelevant values.
% Also, $I_{i,j}$ considers only pixels inside or near the target object, since $X'_{t-1}$ hampers the inflow of irrelevant information which is far from the target object by $\hat{H}_{t-1}$.
This operation is similar to global pooling and region-based operation \cite{caesar2016region} because of making one representative value from the whole $HW$-sized channel and concentrating on a certain region.
For example, if the hexagon in Fig. \ref{fig:template} (a) indicates the estimated location of the target from the previous mask, the information outside of the hexagon is suppressed.
Then $f(X'_{t-1})$ and $g(X'_{t-1})$ are compared with each other along the whole $HW$ plane.
If the two channels are similar, the resultant value of $I$ will be high (red pixel in Fig. \ref{fig:template}(a)); otherwise, it will be low (blue pixel).
Finally, we have $c_{tp}$ embedding vectors of size $1\times c_{tp}$ containing information about the target object.
The final long-term template $TP_{t}$ is updated by weighted-average of the embedding matrix $I$ and the previous template $TP_{t-1}$ as below:
\begin{equation}
\label{eq:tmpUP}
    TP_{t} = \frac{t-1}{t}TP_{t-1} + \frac{1}{t}I.
\end{equation}

The template attention module generates a attention map $A_t \in \mathbb{R}^{c_{tp} \times H\times W}$ by attending on each channel of the query feature map $q(X_t) \in \mathbb{R}^{c_{tp}\times H\times W}$ through the template $TP_{t}$ as follows:
\begin{equation}
\label{eq:tmpOP}
    A_t = TP_{t} \times q(X_t).
\end{equation}
In doing so, the previous estimated mask heatmap $\hat{H}_{t-1}$ is concatenated with the backbone feature map $f8_t$, and the concatenated feature is forwarded to a convolution layer to produce a feature map $X_t$.
Therefore, only certain region of $X_t$ is enhanced by the previous location of target.
% Therefore, the previous location of the target enhances only a certain region of $X_t$.
%  is a feature map where only certain region of around the previous location of the target is enhanced.
Then, $X_t$ is forwarded to several convolution layers to generate a query feature map $q(X_t)$ as shown in Fig. \ref{fig:template}.
In Eq. (\ref{eq:tmpOP}), the attention map $A_t$ is generated by measuring correlations between each row of $TP_t$ (template vector) and each query feature vector from $q(X_t)$, both of which are of a length $c_{tp}$.
When the template vector is highly correlated with the query feature, the resultant $A_t$ value will be high (red pixel in Fig. \ref{fig:template} (a)).
Otherwise, it will be low (blue in Fig. \ref{fig:template} (a)).
After then, the $A_t$ and modified feature map $f8'_t$ are concatenated to make the final long-term similarity map $S_t^L$ by blending both results as shown in the bottom of Fig. \ref{fig:template} (b).

To reduce computational cost while retaining a large receptive field, we use group convolution (group size of 4) with a large kernel size of $5\times5$ for generating $f(\cdot)$, $g(\cdot)$ and $q(\cdot)$. 
While, depth-wise convolutions cost less than the group convolution, we do not use them because their larger group count adversely impacts the model execution time~\cite{ma2018shufflenet}.
We select LeakyReLU as the non-linearity to avoid the dying ReLU problem.
 We empirically determine that using a point-wise convolution first then applying the group convolution achieves better accuracy (shown in Fig. \ref{fig:template} (b)).

\begin{table}[t]
  \centering
  
    \begin{tabular}{l|ccccc}
     \Xhline{3\arrayrulewidth}
          & Read& Seg& Update & \#Param & $J\&F$ \\
          \hline
    GC    & 1.05 G  &  36.8 G     &  37.1 G     &   38 M     & 86.6\\
    Ours  & 0.08 G  &   5.29 G    &   0.06 G    &   1.6 M    & 79.5 \\
     \Xhline{3\arrayrulewidth}
    \end{tabular}%
    \caption{The complexity and accuracy comparison between GC and ours when the input image size is $480\times 853$.
    \textit{Read, Seg}, and \textit{Update} mean the requirement of FLOPS for each operation.
    \textit{Read} indicates reading a memory or a template and dot-producting with query feature map for making $A_t$.
    \textit{Seg} and \textit{Update} denote making a segmentation mask without a decoding stage, and updating a memory or a template.
    Our method reduces lots of computations for updating the template.}
  \label{tab:flop}%
  
\end{table}%

Our template attention module has some similarity to GC but is conceptually very different and computationally much cheaper, as shown in Table~\ref{tab:flop}.
Unlike GC, which is a memory network approach, our method is a kind of template matching approach.
Specifically, GC extracts backbone features again from the new input combining image and mask for generating new memory. 
Then, it produces a global context matrix by different-sized key and value.
However, our template method just combines the current estimated mask and the already calculated backbone feature. 
Then, we use the same-sized feature maps for self-attention to construct multiple embedding vectors representing various characteristics of the target.

\begin{figure*}[t]
\centering
% \fbox{\rule{0pt}{2in} \rule{.9\linewidth}{0pt}}
    \includegraphics[width=0.90\linewidth]{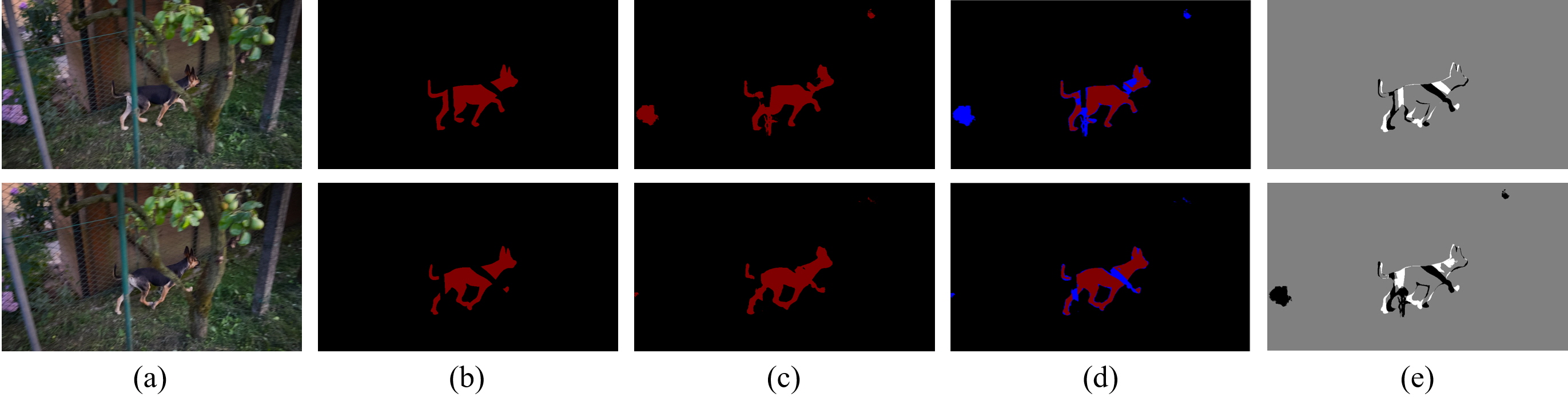}  
   \caption{((a)-(d)) frame $t-1$ and $t$ from top to bottom. (a) Input image. (b) Ground truth. (c) Our result.
   (d) Estimated mask with color marking. Blue color means wrong segmentation result, and the blue region in frame $t$ is corrected from frame $t-1$.  
   (e) Visualizing $\pi_{t,2}$.  Top: $H_t - H_{t-1}$, Bottom: $H_t - \hat{H}_{t-1} $. 
   $H_t - H_{t-1}$ can not remove false positive region in the top of (c).}
\label{fig:TLoss}
\end{figure*}
%%--------------------Fig temporal consistecy loss -------------------------------------

\subsection{Temporal Consistency Loss}
\label{subsec:TCloss}

Our adaptive template deals with the target shape-changing problem by analyzing a backbone feature and an estimated mask along the whole executed frames.
However, using previous estimation incurs the innate error propagation issue.
For example, when the template is updated with a wrong result, this template will gradually lead to incorrect tracking.
However, when the model gets right transition information about how to correct the wrong estimation from the previous frame, the model can mitigate this error propagation problem.
For this reason, we calculate a transition matrix $\hat{\pi}_t$ from $S_t^L$ by a single convolution layer as shown in Fig. \ref{fig:Network}.
We design a novel template consistency loss $L_{tc}$ by $\hat{\pi}_t$.
This loss encourages the model to learn correction power for better consistency across frames:
\begin{equation}
\label{eq:pi}
    \pi_t = H_{t}-\hat{H}_{t-1}, \quad
%\end{equation}
%\begin{equation}
%\label{eq:Ltc}
    L_{tc} = ||\hat{\pi}_t -  \pi_t ||_2^2.
\end{equation}

As a new learning target, we make a target transition matrix from ground truth heatmap $H_{t}$ and previous estimated mask heatmap $\hat{H}_{t-1}$ as in Eq. (\ref{eq:pi}).
Note that the first and the second channel of $H_{t}$ are the probability of background and foreground from a ground truth mask of frame $t$, respectively.
By Eq. (\ref{eq:pi}), the range of $\pi_t$ becomes $(-1,1)$ and $\pi_t$ consists of two channel feature map indicating transition tendency from $t-1$ to $t$.
In detail, the first channel contains transition tendency of the background while the second is for the foreground.
For example, the value of $\pi_{t,2}^{i,j}$ is the $(i,j)$ element of $\pi_t$ in the second channel.
When the value is closer to $1$, it boosts the estimated class to change into foreground from frame $t-1$ to $t$ at position $(i,j)$ .
On the other hand, if the value is close to $-1$, it prevents the estimated class from turning to the foreground.
Finally, when the value is close to $0$, it keeps the same estimated class of frame $t-1$ for frame $t$.

It is important that we use $\hat{H}_{t-1}$ instead of $H_{t-1}$ as illustrated in Fig. \ref{fig:TLoss}. 
Fig. \ref{fig:TLoss}(b) shows ground truth masks, and (c) is the estimated masks at frame $t-1$ (top) and $t$ (bottom).
The first row of Fig. \ref{fig:TLoss}(e) is a visualization of $(H_{t}-{H}_{t-1})$ that can not correct the wrong estimation but maintain the false positive region from the frame $t-1$ to $t$.
However, the second row of Fig. \ref{fig:TLoss}(e) is a visualization of $( H_{t}-\hat{H}_{t-1})$ that guides the estimation to remove false positive region of the frame $t-1$. 
Fig. \ref{fig:TLoss}(d) is marked by blue color for denoting false estimation results comparing between (b) and (c).
As shown in Fig. \ref{fig:TLoss}(d), the transition matrix $\pi_{t,2}$ helps reducing the false positive region from frame $t-1$ to $t$.
With $L_{tc}$, the overall loss becomes:
\begin{equation}
\label{eq:Loss}
Loss = CE(\hat{y}_t, y_t) + \lambda L_{tc},
\end{equation}
where $\lambda$ is a hyper-parameter that makes the balanced scale between the loss terms, and we set $\lambda =5$.
$CE$ denotes the cross entropy between the pixel-wise ground truth $y_t$ at frame $t$ and its predicted value $\hat{y}$.

\begin{table*}[t]
  \centering
 
    \begin{tabular}{l l | cc|ccc|cc|c}
   \Xhline{3\arrayrulewidth}
          &       & \multicolumn{2}{c|}{Model Method} & \multicolumn{3}{c|}{Train Dataset} &  & &  \\
    Method                               & Backbone & OnlineL & Mem\textbackslash Tmp    & YTB & Seg & Synth &   DV17    &     DV16   &  FPS\\
    \hline
    OnAVOS \cite{DBLP:conf/bmvc/VoigtlaenderL17} & VGG16     & o    & -     & - & o & - & 67.9 & 85.5 & 0.08 \\
    OSVOS-S \cite{maninis2018video}              & VGG16     & o    & -     & - & o & - & 68.0 & 86.5 & 0.22 \\
    FRTM-VOS \cite{robinson2020learning}         & ResNet101 & o    & -     & o & - & - & 76.7 & 83.5 & 21.9 \\
    % FRTM-VOS-fast & ResNet18 & o     &   -    & o     &   -    &  -     & 70.2  & 78.5  & 41.3 \\
    \hline
    STM \cite{oh2019video}                       & ResNet50  & -    & o     & o & - & o & 81.8 & 89.3 & 6.25 \\
    GC \cite{li2020fast}                         & ResNet50  & -    & o     & o & - & o & 71.4 & 86.6 & 25.0 \\
     \hline
    OSMN \cite{yang2018efficient}                & VGG16     & -    & o     & - & o & - & 54.8 & 73.5 & 7.69\\
    RANet \cite{wang2019ranet}                   & ResNet101 & -    & o     & - & - & o & 65.7 & 85.5 & 30.3 \\
    A-GAME \cite{johnander2019generative}        & ResNet101 & -    & o     & o & - & o & 70.0 & 82.1 & 14.3 \\
    FEELVOS \cite{voigtlaender2019feelvos}       & Xception 65 & -  & o     & o & o & - & 71.5 & 81.7 & 2.22 \\
    SiamMask \cite{wang2019fast}                 & ResNet50 & -     & o     & o & o & - & 56.4 & 69.8 & 55.0 \\ 
    \hline
    \textbf{TTVOS (Ours)} & \textbf{HRNet}      & \textbf{-} & \textbf{o} & \textbf{o} & \textbf{-} & \textbf{o} & \textbf{58.7} & \textbf{79.5} & \textbf{73.8} \\
    \textbf{TTVOS-RN (Ours)} & \textbf{ResNet50}& \textbf{-} & \textbf{o} & \textbf{o} & \textbf{-} & \textbf{o} & \textbf{67.8} & \textbf{83.8} & \textbf{39.6} \\
    \Xhline{3\arrayrulewidth}
    \end{tabular}%
     \caption{Quantitative comparison on DAVIS benchmark validation set. 
    %  DAVIS 17 is multi-object segmentation and DAVIS 16 is single-object one.
     OnlineL and Mem\textbackslash Tmp  denotes using online-learning method and using memory or template for embedding target information in each network. 
     YTB is using Youtube-VOS for training.
     Seg is segmentation dataset for pre-training by Pascal \cite{everingham2015pascal} or COCO \cite{lin2014microsoft}.
     Synth is using saliency dataset for making synthetic video clip by affine transformation.
     } 
  \label{tab:main}%
\end{table*}%

\begin{figure*}[t]
\centering
% \fbox{\rule{0pt}{2.5in} \rule{.9\linewidth}{0pt}}
    \includegraphics[width=0.88\linewidth]{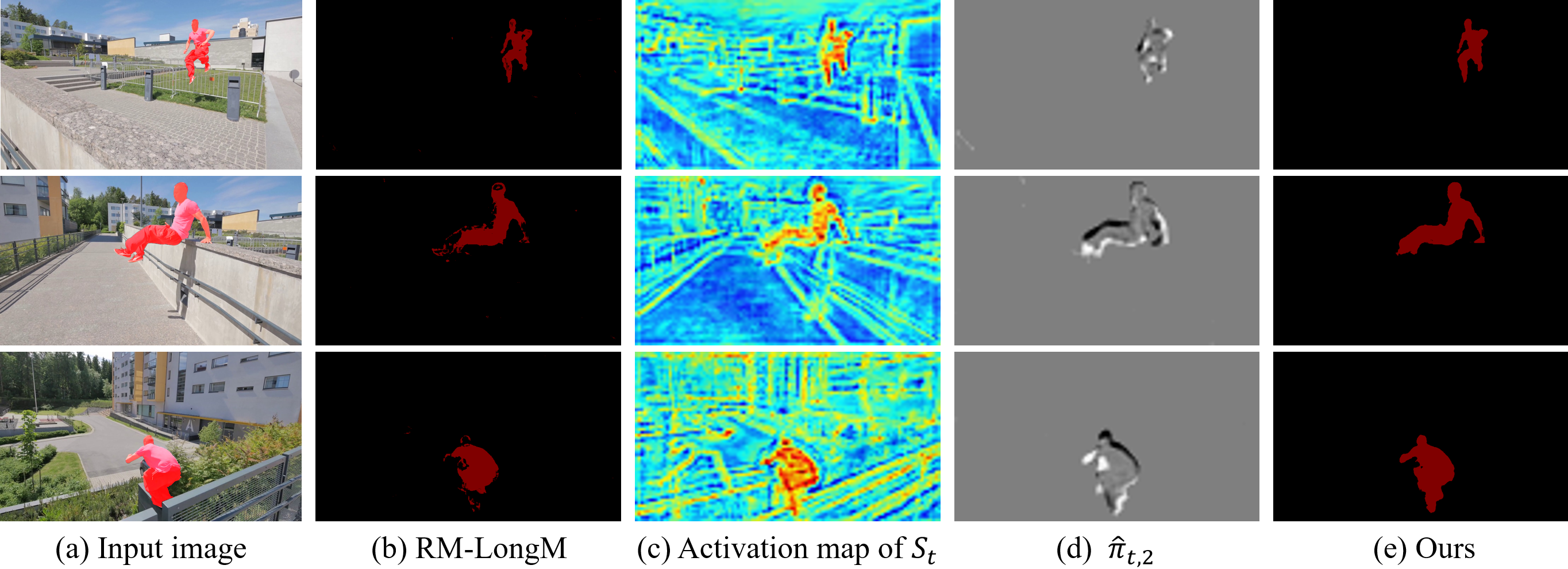}  
   \caption{Example of \textit{parkour} for frame 1, 34 and 84 from top to Bottom.
   Column (a) shows input images overlapped with the ground truth masks.
   RM-LongM denotes estimated results removing long-term matching information by replacing to zeros. }
\label{fig:vis}
\end{figure*}

\section{Experiment}
\label{sec:exp}
Here, we show various evaluations by using DAVIS benchmarks~\cite{Pont-Tuset_arXiv_2017,perazzi2016benchmark}.
DAVIS16 is a single object task consisting of $30$ training videos and $20$ validation videos, and DAVIS17 is a multiple object task with $60$ training videos and $30$ validation videos.
We evaluated our model by using official benchmark code \footnote{https://github.com/davisvideochallenge/davis2017-evaluation}.
The DAVIS benchmark reports model accuracy by average of mean Jaccard index $J$ and mean boundary score $F$.
$J$ index measures overall accuracy by comparing estimated mask and ground truth mask.
$F$ score focuses more contour accuracy by delimiting the spatial extent of the mask.

\noindent
\textbf{Implementation Detail: }
We used HRNetV2-W18-Small-v1 \cite{WangSCJDZLMTWLX19} for a lightweight backbone network and initialized it from the pre-trained parameters from the official code\footnote{https://github.com/HRNet/HRNet-Semantic-Segmentation}.
We froze every backbone layer except the last block.
The size of the smallest feature map is $1/32$ of the input image.
We upsampled the feature map and concatenated it with the second smallest feature map whose size is $1/16$ of the input image.
We used ADAM optimizer for training our model.
First, we pre-trained with synthetic video clip from image dataset, after then we trained with video dataset with single GPU following \cite{oh2019video,voigtlaender2019feelvos,wang2019fast,johnander2019generative}.

\noindent
\textbf{{Pre-train with images}: }
We followed \cite{li2020fast,oh2019video,wang2019ranet} pre-training method, which applies random affine transformation to a static image for generating synthetic video clip.
We used the saliency detection dataset MSRA10K~\cite{cheng2014global}, ECSSD~\cite{yan2013hierarchical}, and HKU-IS~\cite{li2015visual} for various static images.
Synthetic video clips consisting of three frames with a size of $240 \times 432$ were generated.
We trained 100 epochs with an initial learning rate to $1e^{-4}$ and a batch size to $24$.

\noindent
\textbf{{Main-train with videos}: }
We initialized the whole network with the best parameters from the previous step and trained the model to video dataset.
We used a two-stage training method; for the first $100$ epochs, we only used Youtube-VOS with $240 \times 432$ image. We then trained on the DAVIS16 dataset with $480 \times 864$ image for an additional $100$ epochs.
Both training, we used $8$ consecutive frames for a batch, and we set the batch size to $8$ and an initial learning rate to $1e^{-4}$.

\subsection{DAVIS Benchmark Result}
\label{subsec:davis}

\noindent
\textbf{{Comparison to state-of-the-art }: }
We compared our method with other recent models as shown in Table \ref{tab:main}.
We report backbone models and training datasets for clarification because each model has a different setting.
Furthermore, we also show additional results with ResNet50 because some recent models utilized ResNet50 for extracting features.

Our result shows the best accuracy among models with similar speed.
Specifically, SiamMask is one of the popular fast template matching methods, and our model has better accuracy and speed than SiamMask on both DAVIS16 and DAVIS17 benchmark.
When we used ResNet50, our model has better or competitive results with FRTM-VOS, A-GAME, RANet, and FEELVOS.
Also, this ResNet50 based model decreases DAVIS16 accuracy by $2.8\%$ but the speed becomes 1.6 times faster than GC.
Therefore, our method achieves favorable performance among fast VOS models and reduces the performance gap from the online-learning and memory network based models.

\begin{figure}[t]
\centering
    \includegraphics[width=0.98\columnwidth]{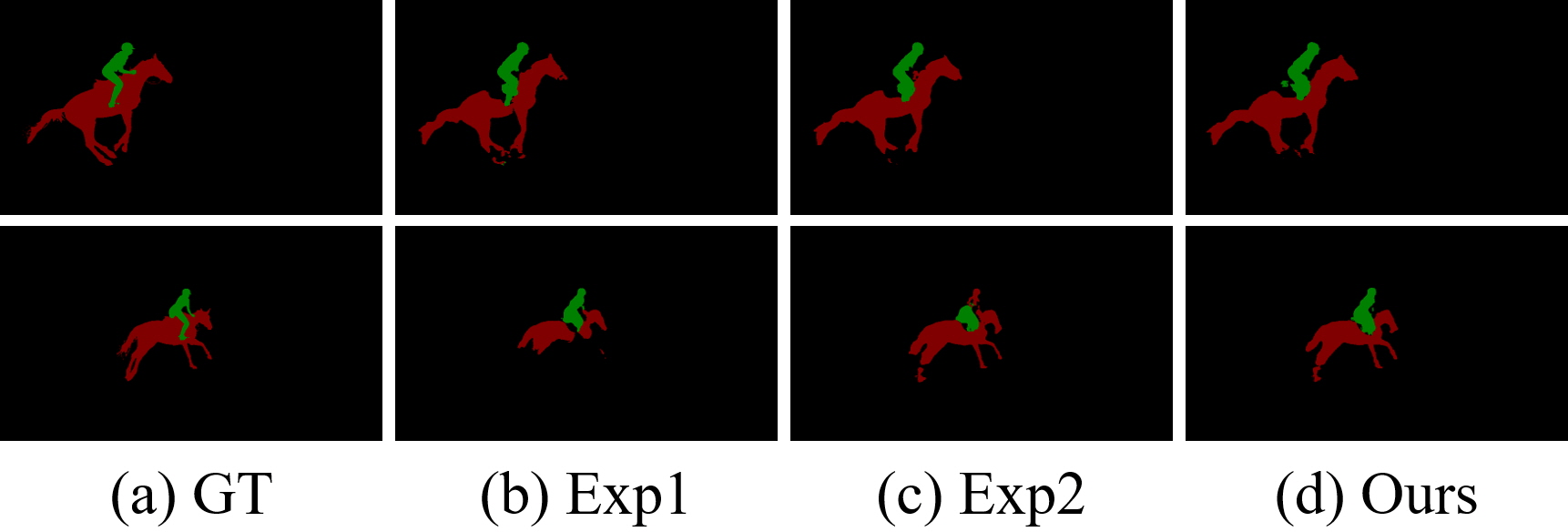}  
   \caption{\textit{Horsejump-high} example of ablation study for frame $3$ and $37$ from top to bottom.
   (a) Ground truth. (b) Using only short-term matching. (c) Using only long-term matching. (d) Our proposed method (Exp6).}
\label{fig:ablation}
\end{figure}

\noindent
\textbf{{Ablation Study }: }
For proving our proposed methods, we performed an ablative analysis on DAVIS16 and DAVIS17 benchmark as shown in Table \ref{tab:ablation}.
SM and LM mean short-term matching and long-term matching, respectively. 
When we do not use short-term matching or long-term matching, we replaced the original matching method into concatenating the previous mask heatmap and the current feature map.
After then the concatenated feature map is forwarded by several convolution layers.
Lup represents updating the long-term template at every frame. 
If not used, the model never updates the template.
TC denotes using temporal consistency loss. 
Without this, the model only uses a cross entropy loss.
M denotes using the original ground truth mask for the initial condition; if M is not checked, a box-shaped mask is used for the initial condition like SiamMask. 
Exp1 is using only short-term matching, and Exp2 is using only long-term matching.
Exp3-6 uses both matching methods.
Table \ref{tab:ablation} is the corresponding accuracy for each ablation experiment, and Fig. \ref{fig:ablation} visualizes efficacy of each template matching.

\begin{table}[t]
  \centering
      \begin{tabular}{c|ccccc|cc}
      \Xhline{3\arrayrulewidth}
    Exp   & SM   & LM  & Lup & TC    & M  & DV17  & DV16 \\
    \hline
    1     & o     & -     & -     & -     & o     & 57.0    & 75.9 \\
    2     & -     & o     & o     & -     & o     & 54.5  & 78.8 \\
    3     & o     & o     & o     & -     & o     & 57.5  & 77.1 \\
    4     & o     & o     & o     & o     & -     & 58.6  & 77.6 \\
    5     & o     & o     & -     & o     & o     & 57.2  & 77.4 \\
    \textbf{6}    & \textbf{o} & \textbf{o} & \textbf{o} & \textbf{o} & \textbf{o} & \textbf{58.7} & \textbf{79.5} \\
    \Xhline{3\arrayrulewidth}
    \end{tabular}%
  
    \caption{Ablation study on DAVIS16 and DAVIS17. SM, LM, TC means short-term matching, long-term matching and temporal consistency loss.
    Lup represents updating long-term template at every frame, and M is using original ground truth mask for initial condition. }
  \label{tab:ablation}%
\end{table}%

We found that short-term matching helps maintain objects ID from localization clue, and long-term matching improves mask quality by enhancing the detailed regions.
For example, Exp1 keeps object ID but fails to make an accurate mask for horse legs, as shown in Fig. \ref{fig:ablation}(b).
On the contrary, Exp2 makes accurate shape but loses green-object (rider) ID as shown in Fig. \ref{fig:ablation}(c).
Exp2 shows performance degradation on multi-object tracking task (DAVIS 17) due to failure in maintaining object ID, even it generates more accurate masks than Exp1.
Therefore, Exp1 achieves better performance in DAVIS17, and Exp2 shows high accuracy in DAVIS16.
Exp3 gets every advantage from both template matching methods, and Fig. \ref{fig:ablation}(d) is our proposed method results (Exp6), which do not lose object ID and generate delicate masks with high performance on both benchmarks.
Exp4-6 explain why our model shows better performance than SiamMask, even using a more lightweight backbone.
The initial condition of the box shape mask does not degrade performance a lot comparing with Exp6.
However, when the model does not update the long-term template, the accuracy degrades a lot from our proposed method.

\noindent
\textbf{{Temporal Consistency Loss }: }
We conducted further experiments for proving the efficacy of our temporal consistency loss with FRTM-VOS,
which is one of the fast online-learning methods, using ResNet101 and ResNet18 for the backbone network.
We implemented our proposed loss function based on FRTM-VOS official code\footnote{https://github.com/andr345/frtm-vos}, and followed their training strategy.
Our proposed loss is more useful in the lightweight backbone network (ResNet18) as shown in Table \ref{tab:TC}.
When we applied our loss to the ResNet101 model, the accuracy on DAVIS17 decreased slightly by 0.1\%, but it increased 1.7\% on DAVIS16.
In the ResNet18 model, we improved the accuracy a lot on both DAVIS17 and DAVIS16.
We conjecture that using our loss not only improves mask quality but also resolves a problem of overfeating due to fine-tuning by a given condition.

\begin{table}[t]
  \centering
      \begin{tabular}{l|c|cc}
      \Xhline{3\arrayrulewidth}
      & Backbone & DV17 & DV16 \\
          \hline
     \multirow{2}{*}{FRTM-VOS~\cite{robinson2020learning}}    & ResNet101 & 76.7  & 83.5 \\
          & ResNet18 & 70.2  & 78.5 \\
          \hline
    \multirow{2}{*}{with TC Loss}    & ResNet101 & 76.6  & 85.2 \\
                                     & ResNet18 & 71.8  & 82.0 \\
          \Xhline{3\arrayrulewidth}
    \end{tabular}%
     \caption{DAVIS17 and DAVIS16 results when additional applying temporal consistency loss (TC Loss).  }
  \label{tab:TC}%
 \end{table}

\section{Conclusion}
\label{sec:conclusion}
Many semi-VOS methods have improved accuracy, but they are hard to utilize in real-world applications due to tremendous complexity.
To resolve this problem, we proposed a novel lightweight semi-VOS model %\nj{based on the} template matching method,
consisting of short-term and long-term matching modules.
%We design\nj{ed} a lightweight semi-VOS model by a short-term and a long-term template matching.
The short-term matching enhances localization, while long-term matching improves mask quality by an adaptive template.
However, using past estimated results incurs an error-propagation problem.
To mitigate this problem, we also devised a new temporal consistency loss to correct false estimated regions by the concept of the transition matrix.
Our model achieves fast inference time while reducing the performance gap from heavy models.
We also showed that the proposed temporal consistency loss can improves accuracy of other models.

% \noindent
% \textbf{Future Work :}
% We improved lots of accuracy in single object task but still \nj{there is} huge performance gap in multi-object task.
% Also, we believe there is a solution that skips several tasks using the previous information for a more efficient model.

\section*{ACKNOWLEDGMENTS}
\label{ack}
This work was supported by the ICT R\&D program of MSIP\slash IITP, Korean Government (2017-0-00306).

\bibliographystyle{ACM-Reference-Format}
\bibliography{egbib}


\begin{thebibliography}{46}

%%% the Web address.

\ifx \showCODEN    \undefined \def \showCODEN     #1{\unskip}     \fi
\ifx \showDOI      \undefined \def \showDOI       #1{#1}\fi
\ifx \showISBNx    \undefined \def \showISBNx     #1{\unskip}     \fi
\ifx \showISBNxiii \undefined \def \showISBNxiii  #1{\unskip}     \fi
\ifx \showISSN     \undefined \def \showISSN      #1{\unskip}     \fi
\ifx \showLCCN     \undefined \def \showLCCN      #1{\unskip}     \fi
\ifx \shownote     \undefined \def \shownote      #1{#1}          \fi
\ifx \showarticletitle \undefined \def \showarticletitle #1{#1}   \fi
\ifx \showURL      \undefined \def \showURL       {\relax}        \fi
% invisible to TeX
\providecommand\bibfield[2]{#2}
\providecommand\bibinfo[2]{#2}
\providecommand\natexlab[1]{#1}
\providecommand\showeprint[2][]{arXiv:#2}

\bibitem[\protect\citeauthoryear{Bertinetto, Valmadre, Henriques, Vedaldi, and
  Torr}{Bertinetto et~al\mbox{.}}{2016}]%
        {bertinetto2016fully}
\bibfield{author}{\bibinfo{person}{Luca Bertinetto}, \bibinfo{person}{Jack
  Valmadre}, \bibinfo{person}{Jo{\~a}o~F Henriques}, \bibinfo{person}{Andrea
  Vedaldi}, {and} \bibinfo{person}{Philip H~S Torr}.}
  \bibinfo{year}{2016}\natexlab{}.
\newblock \showarticletitle{Fully-Convolutional Siamese Networks for Object
  Tracking}. In \bibinfo{booktitle}{\emph{ECCV 2016 Workshops}}.
  \bibinfo{pages}{850--865}.
\newblock


\bibitem[\protect\citeauthoryear{Caelles, Maninis, Pont-Tuset, Leal-Taix\'e,
  Cremers, and {Van Gool}}{Caelles et~al\mbox{.}}{2017}]%
        {Cae+17}
\bibfield{author}{\bibinfo{person}{S. Caelles}, \bibinfo{person}{K.K. Maninis},
  \bibinfo{person}{J. Pont-Tuset}, \bibinfo{person}{L. Leal-Taix\'e},
  \bibinfo{person}{D. Cremers}, {and} \bibinfo{person}{L. {Van Gool}}.}
  \bibinfo{year}{2017}\natexlab{}.
\newblock \showarticletitle{One-Shot Video Object Segmentation}. In
  \bibinfo{booktitle}{\emph{Computer Vision and Pattern Recognition (CVPR)}}.
\newblock


\bibitem[\protect\citeauthoryear{Caesar, Uijlings, and Ferrari}{Caesar
  et~al\mbox{.}}{2016}]%
        {caesar2016region}
\bibfield{author}{\bibinfo{person}{Holger Caesar}, \bibinfo{person}{Jasper
  Uijlings}, {and} \bibinfo{person}{Vittorio Ferrari}.}
  \bibinfo{year}{2016}\natexlab{}.
\newblock \showarticletitle{Region-based semantic segmentation with end-to-end
  training}. In \bibinfo{booktitle}{\emph{European Conference on Computer
  Vision}}. Springer, \bibinfo{pages}{381--397}.
\newblock


\bibitem[\protect\citeauthoryear{Cheng, Tsai, Wang, and Yang}{Cheng
  et~al\mbox{.}}{2017}]%
        {cheng2017segflow}
\bibfield{author}{\bibinfo{person}{Jingchun Cheng}, \bibinfo{person}{Yi-Hsuan
  Tsai}, \bibinfo{person}{Shengjin Wang}, {and} \bibinfo{person}{Ming-Hsuan
  Yang}.} \bibinfo{year}{2017}\natexlab{}.
\newblock \showarticletitle{Segflow: Joint learning for video object
  segmentation and optical flow}. In \bibinfo{booktitle}{\emph{Proceedings of
  the IEEE international conference on computer vision}}.
  \bibinfo{pages}{686--695}.
\newblock


\bibitem[\protect\citeauthoryear{Cheng, Mitra, Huang, Torr, and Hu}{Cheng
  et~al\mbox{.}}{2014}]%
        {cheng2014global}
\bibfield{author}{\bibinfo{person}{Ming-Ming Cheng}, \bibinfo{person}{Niloy~J
  Mitra}, \bibinfo{person}{Xiaolei Huang}, \bibinfo{person}{Philip~HS Torr},
  {and} \bibinfo{person}{Shi-Min Hu}.} \bibinfo{year}{2014}\natexlab{}.
\newblock \showarticletitle{Global contrast based salient region detection}.
\newblock \bibinfo{journal}{\emph{IEEE transactions on pattern analysis and
  machine intelligence}} \bibinfo{volume}{37}, \bibinfo{number}{3}
  (\bibinfo{year}{2014}), \bibinfo{pages}{569--582}.
\newblock


\bibitem[\protect\citeauthoryear{Cho, Cho, Chung, Lee, and Lee}{Cho
  et~al\mbox{.}}{2020}]%
        {cho2020crvos}
\bibfield{author}{\bibinfo{person}{Suhwan Cho}, \bibinfo{person}{MyeongAh Cho},
  \bibinfo{person}{Tae-young Chung}, \bibinfo{person}{Heansung Lee}, {and}
  \bibinfo{person}{Sangyoun Lee}.} \bibinfo{year}{2020}\natexlab{}.
\newblock \showarticletitle{CRVOS: Clue Refining Network for Video Object
  Segmentation}.
\newblock \bibinfo{journal}{\emph{arXiv preprint arXiv:2002.03651}}
  (\bibinfo{year}{2020}).
\newblock


\bibitem[\protect\citeauthoryear{Dutt~Jain, Xiong, and Grauman}{Dutt~Jain
  et~al\mbox{.}}{2017}]%
        {dutt2017fusionseg}
\bibfield{author}{\bibinfo{person}{Suyog Dutt~Jain}, \bibinfo{person}{Bo
  Xiong}, {and} \bibinfo{person}{Kristen Grauman}.}
  \bibinfo{year}{2017}\natexlab{}.
\newblock \showarticletitle{FusionSeg: Learning to combine motion and
  appearance for fully automatic segmentation of generic objects in videos}. In
  \bibinfo{booktitle}{\emph{Proceedings of the IEEE conference on computer
  vision and pattern recognition}}. \bibinfo{pages}{3664--3673}.
\newblock


\bibitem[\protect\citeauthoryear{Everingham, Eslami, Van~Gool, Williams, Winn,
  and Zisserman}{Everingham et~al\mbox{.}}{2015}]%
        {everingham2015pascal}
\bibfield{author}{\bibinfo{person}{Mark Everingham}, \bibinfo{person}{SM~Ali
  Eslami}, \bibinfo{person}{Luc Van~Gool}, \bibinfo{person}{Christopher~KI
  Williams}, \bibinfo{person}{John Winn}, {and} \bibinfo{person}{Andrew
  Zisserman}.} \bibinfo{year}{2015}\natexlab{}.
\newblock \showarticletitle{The pascal visual object classes challenge: A
  retrospective}.
\newblock \bibinfo{journal}{\emph{International journal of computer vision}}
  \bibinfo{volume}{111}, \bibinfo{number}{1} (\bibinfo{year}{2015}),
  \bibinfo{pages}{98--136}.
\newblock


\bibitem[\protect\citeauthoryear{Hu, Wang, Kong, Kuen, and Tan}{Hu
  et~al\mbox{.}}{2018}]%
        {hu2018motion}
\bibfield{author}{\bibinfo{person}{Ping Hu}, \bibinfo{person}{Gang Wang},
  \bibinfo{person}{Xiangfei Kong}, \bibinfo{person}{Jason Kuen}, {and}
  \bibinfo{person}{Yap-Peng Tan}.} \bibinfo{year}{2018}\natexlab{}.
\newblock \showarticletitle{Motion-guided cascaded refinement network for video
  object segmentation}. In \bibinfo{booktitle}{\emph{Proceedings of the IEEE
  Conference on Computer Vision and Pattern Recognition}}.
  \bibinfo{pages}{1400--1409}.
\newblock


\bibitem[\protect\citeauthoryear{Johnander, Danelljan, Brissman, Khan, and
  Felsberg}{Johnander et~al\mbox{.}}{2019}]%
        {johnander2019generative}
\bibfield{author}{\bibinfo{person}{Joakim Johnander}, \bibinfo{person}{Martin
  Danelljan}, \bibinfo{person}{Emil Brissman}, \bibinfo{person}{Fahad~Shahbaz
  Khan}, {and} \bibinfo{person}{Michael Felsberg}.}
  \bibinfo{year}{2019}\natexlab{}.
\newblock \showarticletitle{A generative appearance model for end-to-end video
  object segmentation}. In \bibinfo{booktitle}{\emph{Proceedings of the IEEE
  Conference on Computer Vision and Pattern Recognition}}.
  \bibinfo{pages}{8953--8962}.
\newblock


\bibitem[\protect\citeauthoryear{Khoreva, Benenson, Ilg, Brox, and
  Schiele}{Khoreva et~al\mbox{.}}{2017}]%
        {khoreva2017lucid}
\bibfield{author}{\bibinfo{person}{Anna Khoreva}, \bibinfo{person}{Rodrigo
  Benenson}, \bibinfo{person}{Eddy Ilg}, \bibinfo{person}{Thomas Brox}, {and}
  \bibinfo{person}{Bernt Schiele}.} \bibinfo{year}{2017}\natexlab{}.
\newblock \showarticletitle{Lucid data dreaming for object tracking}. In
  \bibinfo{booktitle}{\emph{The DAVIS Challenge on Video Object Segmentation}}.
\newblock


\bibitem[\protect\citeauthoryear{Kim, Ma, Kim, Kim, and Yoo}{Kim
  et~al\mbox{.}}{2019}]%
        {kim2019progressive}
\bibfield{author}{\bibinfo{person}{Junyeong Kim}, \bibinfo{person}{Minuk Ma},
  \bibinfo{person}{Kyungsu Kim}, \bibinfo{person}{Sungjin Kim}, {and}
  \bibinfo{person}{Chang~D Yoo}.} \bibinfo{year}{2019}\natexlab{}.
\newblock \showarticletitle{Progressive attention memory network for movie
  story question answering}. In \bibinfo{booktitle}{\emph{Proceedings of the
  IEEE Conference on Computer Vision and Pattern Recognition}}.
  \bibinfo{pages}{8337--8346}.
\newblock


\bibitem[\protect\citeauthoryear{Kivinen, Smola, and Williamson}{Kivinen
  et~al\mbox{.}}{2004}]%
        {kivinen2004online}
\bibfield{author}{\bibinfo{person}{Jyrki Kivinen}, \bibinfo{person}{Alexander~J
  Smola}, {and} \bibinfo{person}{Robert~C Williamson}.}
  \bibinfo{year}{2004}\natexlab{}.
\newblock \showarticletitle{Online learning with kernels}.
\newblock \bibinfo{journal}{\emph{IEEE transactions on signal processing}}
  \bibinfo{volume}{52}, \bibinfo{number}{8} (\bibinfo{year}{2004}),
  \bibinfo{pages}{2165--2176}.
\newblock


\bibitem[\protect\citeauthoryear{Li and Yu}{Li and Yu}{2015}]%
        {li2015visual}
\bibfield{author}{\bibinfo{person}{Guanbin Li} {and} \bibinfo{person}{Yizhou
  Yu}.} \bibinfo{year}{2015}\natexlab{}.
\newblock \showarticletitle{Visual saliency based on multiscale deep features}.
  In \bibinfo{booktitle}{\emph{Proceedings of the IEEE conference on computer
  vision and pattern recognition}}. \bibinfo{pages}{5455--5463}.
\newblock


\bibitem[\protect\citeauthoryear{Li, Shen, and Shan}{Li et~al\mbox{.}}{2020}]%
        {li2020fast}
\bibfield{author}{\bibinfo{person}{Yu Li}, \bibinfo{person}{Zhuoran Shen},
  {and} \bibinfo{person}{Ying Shan}.} \bibinfo{year}{2020}\natexlab{}.
\newblock \showarticletitle{Fast Video Object Segmentation using the Global
  Context Module}. In \bibinfo{booktitle}{\emph{The European Conference on
  Computer Vision (ECCV)}}.
\newblock


\bibitem[\protect\citeauthoryear{Lin, Chou, and Martinez}{Lin
  et~al\mbox{.}}{2020}]%
        {lin2020flow}
\bibfield{author}{\bibinfo{person}{Fanqing Lin}, \bibinfo{person}{Yao Chou},
  {and} \bibinfo{person}{Tony Martinez}.} \bibinfo{year}{2020}\natexlab{}.
\newblock \showarticletitle{Flow Adaptive Video Object Segmentation}.
\newblock \bibinfo{journal}{\emph{Image and Vision Computing}}
  \bibinfo{volume}{94} (\bibinfo{year}{2020}), \bibinfo{pages}{103864}.
\newblock


\bibitem[\protect\citeauthoryear{Lin, Maire, Belongie, Hays, Perona, Ramanan,
  Doll{\'a}r, and Zitnick}{Lin et~al\mbox{.}}{2014}]%
        {lin2014microsoft}
\bibfield{author}{\bibinfo{person}{Tsung-Yi Lin}, \bibinfo{person}{Michael
  Maire}, \bibinfo{person}{Serge Belongie}, \bibinfo{person}{James Hays},
  \bibinfo{person}{Pietro Perona}, \bibinfo{person}{Deva Ramanan},
  \bibinfo{person}{Piotr Doll{\'a}r}, {and} \bibinfo{person}{C~Lawrence
  Zitnick}.} \bibinfo{year}{2014}\natexlab{}.
\newblock \showarticletitle{Microsoft coco: Common objects in context}. In
  \bibinfo{booktitle}{\emph{European conference on computer vision}}. Springer,
  \bibinfo{pages}{740--755}.
\newblock


\bibitem[\protect\citeauthoryear{Ma, Zhang, Zheng, and Sun}{Ma
  et~al\mbox{.}}{2018}]%
        {ma2018shufflenet}
\bibfield{author}{\bibinfo{person}{Ningning Ma}, \bibinfo{person}{Xiangyu
  Zhang}, \bibinfo{person}{Hai-Tao Zheng}, {and} \bibinfo{person}{Jian Sun}.}
  \bibinfo{year}{2018}\natexlab{}.
\newblock \showarticletitle{Shufflenet v2: Practical guidelines for efficient
  cnn architecture design}. In \bibinfo{booktitle}{\emph{Proceedings of the
  European conference on computer vision (ECCV)}}. \bibinfo{pages}{116--131}.
\newblock


\bibitem[\protect\citeauthoryear{Maninis, Caelles, Chen, Pont-Tuset,
  Leal-Taix{\'e}, Cremers, and Van~Gool}{Maninis et~al\mbox{.}}{2018}]%
        {maninis2018video}
\bibfield{author}{\bibinfo{person}{K-K Maninis}, \bibinfo{person}{Sergi
  Caelles}, \bibinfo{person}{Yuhua Chen}, \bibinfo{person}{Jordi Pont-Tuset},
  \bibinfo{person}{Laura Leal-Taix{\'e}}, \bibinfo{person}{Daniel Cremers},
  {and} \bibinfo{person}{Luc Van~Gool}.} \bibinfo{year}{2018}\natexlab{}.
\newblock \showarticletitle{Video object segmentation without temporal
  information}.
\newblock \bibinfo{journal}{\emph{IEEE transactions on pattern analysis and
  machine intelligence}} \bibinfo{volume}{41}, \bibinfo{number}{6}
  (\bibinfo{year}{2018}), \bibinfo{pages}{1515--1530}.
\newblock


\bibitem[\protect\citeauthoryear{Miyato, Maeda, Koyama, and Ishii}{Miyato
  et~al\mbox{.}}{2018}]%
        {miyato2018virtual}
\bibfield{author}{\bibinfo{person}{Takeru Miyato}, \bibinfo{person}{Shin-ichi
  Maeda}, \bibinfo{person}{Masanori Koyama}, {and} \bibinfo{person}{Shin
  Ishii}.} \bibinfo{year}{2018}\natexlab{}.
\newblock \showarticletitle{Virtual adversarial training: a regularization
  method for supervised and semi-supervised learning}.
\newblock \bibinfo{journal}{\emph{IEEE transactions on pattern analysis and
  machine intelligence}} \bibinfo{volume}{41}, \bibinfo{number}{8}
  (\bibinfo{year}{2018}), \bibinfo{pages}{1979--1993}.
\newblock


\bibitem[\protect\citeauthoryear{Oh, Lee, Xu, and Kim}{Oh
  et~al\mbox{.}}{2019}]%
        {oh2019video}
\bibfield{author}{\bibinfo{person}{Seoung~Wug Oh}, \bibinfo{person}{Joon-Young
  Lee}, \bibinfo{person}{Ning Xu}, {and} \bibinfo{person}{Seon~Joo Kim}.}
  \bibinfo{year}{2019}\natexlab{}.
\newblock \showarticletitle{Video object segmentation using space-time memory
  networks}. In \bibinfo{booktitle}{\emph{Proceedings of the IEEE International
  Conference on Computer Vision}}. \bibinfo{pages}{9226--9235}.
\newblock


\bibitem[\protect\citeauthoryear{Perazzi, Khoreva, Benenson, Schiele, and
  Sorkine-Hornung}{Perazzi et~al\mbox{.}}{2017}]%
        {perazzi2017learning}
\bibfield{author}{\bibinfo{person}{Federico Perazzi}, \bibinfo{person}{Anna
  Khoreva}, \bibinfo{person}{Rodrigo Benenson}, \bibinfo{person}{Bernt
  Schiele}, {and} \bibinfo{person}{Alexander Sorkine-Hornung}.}
  \bibinfo{year}{2017}\natexlab{}.
\newblock \showarticletitle{Learning video object segmentation from static
  images}. In \bibinfo{booktitle}{\emph{Proceedings of the IEEE conference on
  computer vision and pattern recognition}}. \bibinfo{pages}{2663--2672}.
\newblock


\bibitem[\protect\citeauthoryear{Perazzi, Pont-Tuset, McWilliams, Van~Gool,
  Gross, and Sorkine-Hornung}{Perazzi et~al\mbox{.}}{2016}]%
        {perazzi2016benchmark}
\bibfield{author}{\bibinfo{person}{Federico Perazzi}, \bibinfo{person}{Jordi
  Pont-Tuset}, \bibinfo{person}{Brian McWilliams}, \bibinfo{person}{Luc
  Van~Gool}, \bibinfo{person}{Markus Gross}, {and} \bibinfo{person}{Alexander
  Sorkine-Hornung}.} \bibinfo{year}{2016}\natexlab{}.
\newblock \showarticletitle{A benchmark dataset and evaluation methodology for
  video object segmentation}. In \bibinfo{booktitle}{\emph{Proceedings of the
  IEEE Conference on Computer Vision and Pattern Recognition}}.
  \bibinfo{pages}{724--732}.
\newblock


\bibitem[\protect\citeauthoryear{Pont-Tuset, Perazzi, Caelles, Arbel\'aez,
  Sorkine-Hornung, and {Van Gool}}{Pont-Tuset et~al\mbox{.}}{2017}]%
        {Pont-Tuset_arXiv_2017}
\bibfield{author}{\bibinfo{person}{Jordi Pont-Tuset}, \bibinfo{person}{Federico
  Perazzi}, \bibinfo{person}{Sergi Caelles}, \bibinfo{person}{Pablo
  Arbel\'aez}, \bibinfo{person}{Alexander Sorkine-Hornung}, {and}
  \bibinfo{person}{Luc {Van Gool}}.} \bibinfo{year}{2017}\natexlab{}.
\newblock \showarticletitle{The 2017 DAVIS Challenge on Video Object
  Segmentation}.
\newblock \bibinfo{journal}{\emph{arXiv:1704.00675}} (\bibinfo{year}{2017}).
\newblock


\bibitem[\protect\citeauthoryear{Robinson, Lawin, Danelljan, Khan, and
  Felsberg}{Robinson et~al\mbox{.}}{2020}]%
        {robinson2020learning}
\bibfield{author}{\bibinfo{person}{Andreas Robinson},
  \bibinfo{person}{Felix~Jaremo Lawin}, \bibinfo{person}{Martin Danelljan},
  \bibinfo{person}{Fahad~Shahbaz Khan}, {and} \bibinfo{person}{Michael
  Felsberg}.} \bibinfo{year}{2020}\natexlab{}.
\newblock \showarticletitle{Learning Fast and Robust Target Models for Video
  Object Segmentation}. In \bibinfo{booktitle}{\emph{Proceedings of the
  IEEE/CVF Conference on Computer Vision and Pattern Recognition}}.
  \bibinfo{pages}{7406--7415}.
\newblock


\bibitem[\protect\citeauthoryear{Sahoo, Pham, Lu, and Hoi}{Sahoo
  et~al\mbox{.}}{2018}]%
        {ijcai2018-369}
\bibfield{author}{\bibinfo{person}{Doyen Sahoo}, \bibinfo{person}{Quang Pham},
  \bibinfo{person}{Jing Lu}, {and} \bibinfo{person}{Steven C.~H. Hoi}.}
  \bibinfo{year}{2018}\natexlab{}.
\newblock \showarticletitle{Online Deep Learning: Learning Deep Neural Networks
  on the Fly}. In \bibinfo{booktitle}{\emph{Proceedings of the Twenty-Seventh
  International Joint Conference on Artificial Intelligence, {IJCAI-18}}}.
  \bibinfo{publisher}{International Joint Conferences on Artificial
  Intelligence Organization}, \bibinfo{pages}{2660--2666}.
\newblock
\urldef\tempurl%
\url{https://doi.org/10.24963/ijcai.2018/369}
\showDOI{\tempurl}


\bibitem[\protect\citeauthoryear{Sevilla-Lara, Sun, Jampani, and
  Black}{Sevilla-Lara et~al\mbox{.}}{2016}]%
        {sevilla2016optical}
\bibfield{author}{\bibinfo{person}{Laura Sevilla-Lara}, \bibinfo{person}{Deqing
  Sun}, \bibinfo{person}{Varun Jampani}, {and} \bibinfo{person}{Michael~J
  Black}.} \bibinfo{year}{2016}\natexlab{}.
\newblock \showarticletitle{Optical flow with semantic segmentation and
  localized layers}. In \bibinfo{booktitle}{\emph{Proceedings of the IEEE
  Conference on Computer Vision and Pattern Recognition}}.
  \bibinfo{pages}{3889--3898}.
\newblock


\bibitem[\protect\citeauthoryear{Shi, Caballero, Husz{\'a}r, Totz, Aitken,
  Bishop, Rueckert, and Wang}{Shi et~al\mbox{.}}{2016}]%
        {shi2016real}
\bibfield{author}{\bibinfo{person}{Wenzhe Shi}, \bibinfo{person}{Jose
  Caballero}, \bibinfo{person}{Ferenc Husz{\'a}r}, \bibinfo{person}{Johannes
  Totz}, \bibinfo{person}{Andrew~P Aitken}, \bibinfo{person}{Rob Bishop},
  \bibinfo{person}{Daniel Rueckert}, {and} \bibinfo{person}{Zehan Wang}.}
  \bibinfo{year}{2016}\natexlab{}.
\newblock \showarticletitle{Real-time single image and video super-resolution
  using an efficient sub-pixel convolutional neural network}. In
  \bibinfo{booktitle}{\emph{Proceedings of the IEEE conference on computer
  vision and pattern recognition}}. \bibinfo{pages}{1874--1883}.
\newblock


\bibitem[\protect\citeauthoryear{Sukhbaatar, Weston, Fergus,
  et~al\mbox{.}}{Sukhbaatar et~al\mbox{.}}{2015}]%
        {sukhbaatar2015end}
\bibfield{author}{\bibinfo{person}{Sainbayar Sukhbaatar},
  \bibinfo{person}{Jason Weston}, \bibinfo{person}{Rob Fergus},
  {et~al\mbox{.}}} \bibinfo{year}{2015}\natexlab{}.
\newblock \showarticletitle{End-to-end memory networks}. In
  \bibinfo{booktitle}{\emph{Advances in neural information processing
  systems}}. \bibinfo{pages}{2440--2448}.
\newblock


\bibitem[\protect\citeauthoryear{Tsai, Yang, and Black}{Tsai
  et~al\mbox{.}}{2016a}]%
        {tsai2016video}
\bibfield{author}{\bibinfo{person}{Yi-Hsuan Tsai}, \bibinfo{person}{Ming-Hsuan
  Yang}, {and} \bibinfo{person}{Michael~J Black}.}
  \bibinfo{year}{2016}\natexlab{a}.
\newblock \showarticletitle{Video segmentation via object flow}. In
  \bibinfo{booktitle}{\emph{Proceedings of the IEEE conference on computer
  vision and pattern recognition}}. \bibinfo{pages}{3899--3908}.
\newblock


\bibitem[\protect\citeauthoryear{Tsai, Yang, and Black}{Tsai
  et~al\mbox{.}}{2016b}]%
        {Tsai_2016_CVPR}
\bibfield{author}{\bibinfo{person}{Yi-Hsuan Tsai}, \bibinfo{person}{Ming-Hsuan
  Yang}, {and} \bibinfo{person}{Michael~J. Black}.}
  \bibinfo{year}{2016}\natexlab{b}.
\newblock \showarticletitle{Video Segmentation via Object Flow}. In
  \bibinfo{booktitle}{\emph{Proceedings of the IEEE Conference on Computer
  Vision and Pattern Recognition (CVPR)}}.
\newblock


\bibitem[\protect\citeauthoryear{Voigtlaender, Chai, Schroff, Adam, Leibe, and
  Chen}{Voigtlaender et~al\mbox{.}}{2019}]%
        {voigtlaender2019feelvos}
\bibfield{author}{\bibinfo{person}{Paul Voigtlaender}, \bibinfo{person}{Yuning
  Chai}, \bibinfo{person}{Florian Schroff}, \bibinfo{person}{Hartwig Adam},
  \bibinfo{person}{Bastian Leibe}, {and} \bibinfo{person}{Liang-Chieh Chen}.}
  \bibinfo{year}{2019}\natexlab{}.
\newblock \showarticletitle{Feelvos: Fast end-to-end embedding learning for
  video object segmentation}. In \bibinfo{booktitle}{\emph{Proceedings of the
  IEEE Conference on Computer Vision and Pattern Recognition}}.
  \bibinfo{pages}{9481--9490}.
\newblock


\bibitem[\protect\citeauthoryear{Voigtlaender and Leibe}{Voigtlaender and
  Leibe}{2017}]%
        {DBLP:conf/bmvc/VoigtlaenderL17}
\bibfield{author}{\bibinfo{person}{Paul Voigtlaender} {and}
  \bibinfo{person}{Bastian Leibe}.} \bibinfo{year}{2017}\natexlab{}.
\newblock \showarticletitle{Online Adaptation of Convolutional Neural Networks
  for Video Object Segmentation}. In \bibinfo{booktitle}{\emph{British Machine
  Vision Conference 2017, {BMVC} 2017, London, UK, September 4-7, 2017}}.
  \bibinfo{publisher}{{BMVA} Press}.
\newblock


\bibitem[\protect\citeauthoryear{Volz, Bruhn, Valgaerts, and Zimmer}{Volz
  et~al\mbox{.}}{2011}]%
        {volz2011modeling}
\bibfield{author}{\bibinfo{person}{Sebastian Volz}, \bibinfo{person}{Andres
  Bruhn}, \bibinfo{person}{Levi Valgaerts}, {and} \bibinfo{person}{Henning
  Zimmer}.} \bibinfo{year}{2011}\natexlab{}.
\newblock \showarticletitle{Modeling temporal coherence for optical flow}. In
  \bibinfo{booktitle}{\emph{2011 International Conference on Computer Vision}}.
  IEEE, \bibinfo{pages}{1116--1123}.
\newblock


\bibitem[\protect\citeauthoryear{Wang, Sun, Cheng, Jiang, Deng, Zhao, Liu, Mu,
  Tan, Wang, Liu, and Xiao}{Wang et~al\mbox{.}}{2019a}]%
        {WangSCJDZLMTWLX19}
\bibfield{author}{\bibinfo{person}{Jingdong Wang}, \bibinfo{person}{Ke Sun},
  \bibinfo{person}{Tianheng Cheng}, \bibinfo{person}{Borui Jiang},
  \bibinfo{person}{Chaorui Deng}, \bibinfo{person}{Yang Zhao},
  \bibinfo{person}{Dong Liu}, \bibinfo{person}{Yadong Mu},
  \bibinfo{person}{Mingkui Tan}, \bibinfo{person}{Xinggang Wang},
  \bibinfo{person}{Wenyu Liu}, {and} \bibinfo{person}{Bin Xiao}.}
  \bibinfo{year}{2019}\natexlab{a}.
\newblock \showarticletitle{Deep High-Resolution Representation Learning for
  Visual Recognition}.
\newblock \bibinfo{journal}{\emph{TPAMI}} (\bibinfo{year}{2019}).
\newblock


\bibitem[\protect\citeauthoryear{Wang, Zhang, Bertinetto, Hu, and Torr}{Wang
  et~al\mbox{.}}{2019c}]%
        {wang2019fast}
\bibfield{author}{\bibinfo{person}{Qiang Wang}, \bibinfo{person}{Li Zhang},
  \bibinfo{person}{Luca Bertinetto}, \bibinfo{person}{Weiming Hu}, {and}
  \bibinfo{person}{Philip~HS Torr}.} \bibinfo{year}{2019}\natexlab{c}.
\newblock \showarticletitle{Fast online object tracking and segmentation: A
  unifying approach}. In \bibinfo{booktitle}{\emph{Proceedings of the IEEE
  conference on computer vision and pattern recognition}}.
  \bibinfo{pages}{1328--1338}.
\newblock


\bibitem[\protect\citeauthoryear{Wang, Shen, Porikli, and Yang}{Wang
  et~al\mbox{.}}{2018b}]%
        {wang2018semi}
\bibfield{author}{\bibinfo{person}{Wenguan Wang}, \bibinfo{person}{Jianbing
  Shen}, \bibinfo{person}{Fatih Porikli}, {and} \bibinfo{person}{Ruigang
  Yang}.} \bibinfo{year}{2018}\natexlab{b}.
\newblock \showarticletitle{Semi-supervised video object segmentation with
  super-trajectories}.
\newblock \bibinfo{journal}{\emph{IEEE transactions on pattern analysis and
  machine intelligence}} \bibinfo{volume}{41}, \bibinfo{number}{4}
  (\bibinfo{year}{2018}), \bibinfo{pages}{985--998}.
\newblock


\bibitem[\protect\citeauthoryear{Wang, Girshick, Gupta, and He}{Wang
  et~al\mbox{.}}{2018a}]%
        {wang2018non}
\bibfield{author}{\bibinfo{person}{Xiaolong Wang}, \bibinfo{person}{Ross
  Girshick}, \bibinfo{person}{Abhinav Gupta}, {and} \bibinfo{person}{Kaiming
  He}.} \bibinfo{year}{2018}\natexlab{a}.
\newblock \showarticletitle{Non-local neural networks}. In
  \bibinfo{booktitle}{\emph{Proceedings of the IEEE conference on computer
  vision and pattern recognition}}. \bibinfo{pages}{7794--7803}.
\newblock


\bibitem[\protect\citeauthoryear{Wang, Xu, Liu, Zhu, and Shao}{Wang
  et~al\mbox{.}}{2019b}]%
        {wang2019ranet}
\bibfield{author}{\bibinfo{person}{Ziqin Wang}, \bibinfo{person}{Jun Xu},
  \bibinfo{person}{Li Liu}, \bibinfo{person}{Fan Zhu}, {and}
  \bibinfo{person}{Ling Shao}.} \bibinfo{year}{2019}\natexlab{b}.
\newblock \showarticletitle{Ranet: Ranking attention network for fast video
  object segmentation}. In \bibinfo{booktitle}{\emph{Proceedings of the IEEE
  international conference on computer vision}}. \bibinfo{pages}{3978--3987}.
\newblock


\bibitem[\protect\citeauthoryear{Weickert and Schn{\"o}rr}{Weickert and
  Schn{\"o}rr}{2001}]%
        {weickert2001variational}
\bibfield{author}{\bibinfo{person}{Joachim Weickert} {and}
  \bibinfo{person}{Christoph Schn{\"o}rr}.} \bibinfo{year}{2001}\natexlab{}.
\newblock \showarticletitle{Variational optic flow computation with a
  spatio-temporal smoothness constraint}.
\newblock \bibinfo{journal}{\emph{Journal of mathematical imaging and vision}}
  \bibinfo{volume}{14}, \bibinfo{number}{3} (\bibinfo{year}{2001}),
  \bibinfo{pages}{245--255}.
\newblock


\bibitem[\protect\citeauthoryear{Weston, Chopra, and Bordes}{Weston
  et~al\mbox{.}}{2014}]%
        {weston2014memory}
\bibfield{author}{\bibinfo{person}{Jason Weston}, \bibinfo{person}{Sumit
  Chopra}, {and} \bibinfo{person}{Antoine Bordes}.}
  \bibinfo{year}{2014}\natexlab{}.
\newblock \showarticletitle{Memory networks}.
\newblock \bibinfo{journal}{\emph{arXiv preprint arXiv:1410.3916}}
  (\bibinfo{year}{2014}).
\newblock


\bibitem[\protect\citeauthoryear{Yan, Xu, Shi, and Jia}{Yan
  et~al\mbox{.}}{2013}]%
        {yan2013hierarchical}
\bibfield{author}{\bibinfo{person}{Qiong Yan}, \bibinfo{person}{Li Xu},
  \bibinfo{person}{Jianping Shi}, {and} \bibinfo{person}{Jiaya Jia}.}
  \bibinfo{year}{2013}\natexlab{}.
\newblock \showarticletitle{Hierarchical saliency detection}. In
  \bibinfo{booktitle}{\emph{Proceedings of the IEEE conference on computer
  vision and pattern recognition}}. \bibinfo{pages}{1155--1162}.
\newblock


\bibitem[\protect\citeauthoryear{Yang, Wang, Xiong, Yang, and Katsaggelos}{Yang
  et~al\mbox{.}}{2018}]%
        {yang2018efficient}
\bibfield{author}{\bibinfo{person}{Linjie Yang}, \bibinfo{person}{Yanran Wang},
  \bibinfo{person}{Xuehan Xiong}, \bibinfo{person}{Jianchao Yang}, {and}
  \bibinfo{person}{Aggelos~K Katsaggelos}.} \bibinfo{year}{2018}\natexlab{}.
\newblock \showarticletitle{Efficient video object segmentation via network
  modulation}. In \bibinfo{booktitle}{\emph{Proceedings of the IEEE Conference
  on Computer Vision and Pattern Recognition}}. \bibinfo{pages}{6499--6507}.
\newblock


\bibitem[\protect\citeauthoryear{Zhou, Sohn, and Lee}{Zhou
  et~al\mbox{.}}{2012}]%
        {zhou2012online}
\bibfield{author}{\bibinfo{person}{Guanyu Zhou}, \bibinfo{person}{Kihyuk Sohn},
  {and} \bibinfo{person}{Honglak Lee}.} \bibinfo{year}{2012}\natexlab{}.
\newblock \showarticletitle{Online incremental feature learning with denoising
  autoencoders}. In \bibinfo{booktitle}{\emph{Artificial intelligence and
  statistics}}. \bibinfo{pages}{1453--1461}.
\newblock


\bibitem[\protect\citeauthoryear{Zhu, Park, Isola, and Efros}{Zhu
  et~al\mbox{.}}{2017}]%
        {zhu2017unpaired}
\bibfield{author}{\bibinfo{person}{Jun-Yan Zhu}, \bibinfo{person}{Taesung
  Park}, \bibinfo{person}{Phillip Isola}, {and} \bibinfo{person}{Alexei~A
  Efros}.} \bibinfo{year}{2017}\natexlab{}.
\newblock \showarticletitle{Unpaired image-to-image translation using
  cycle-consistent adversarial networks}. In
  \bibinfo{booktitle}{\emph{Proceedings of the IEEE international conference on
  computer vision}}. \bibinfo{pages}{2223--2232}.
\newblock


\bibitem[\protect\citeauthoryear{Zhu, Xu, Bai, Huang, and Bai}{Zhu
  et~al\mbox{.}}{2019}]%
        {zhu2019asymmetric}
\bibfield{author}{\bibinfo{person}{Zhen Zhu}, \bibinfo{person}{Mengde Xu},
  \bibinfo{person}{Song Bai}, \bibinfo{person}{Tengteng Huang}, {and}
  \bibinfo{person}{Xiang Bai}.} \bibinfo{year}{2019}\natexlab{}.
\newblock \showarticletitle{Asymmetric non-local neural networks for semantic
  segmentation}. In \bibinfo{booktitle}{\emph{Proceedings of the IEEE
  International Conference on Computer Vision}}. \bibinfo{pages}{593--602}.
\newblock


\end{thebibliography}

\end{document}